\documentclass{article}
\usepackage{ijcai22}

\usepackage{times}

\usepackage{soul}
\usepackage{url}
\usepackage[hidelinks]{hyperref}
\usepackage[utf8]{inputenc}
\usepackage[small]{caption}
\usepackage{graphicx}
\usepackage{amsmath}
\usepackage{booktabs}
\usepackage{comment}
\usepackage[pagewise]{lineno}

\title{Multi-Exposure HDR Composition by Gated Swin Transformer}

\author{
Rui Zhou
\and
Yan Niu
\affiliations
Jilin University
}

\begin{document}

\maketitle

\begin{abstract}
Fusing a sequence of perfectly aligned images captured at various exposures, has shown great potential to approach High Dynamic Range (HDR) imaging by sensors with limited dynamic range. However, in the presence of large motion of scene objects or the camera, mis-alignment is almost inevitable and leads to the notorious ``ghost''  artifacts. Besides, factors such as the noise in the dark region or color saturation in the over-bright region may also fail to fill local image details to the HDR image. This paper provides a novel multi-exposure fusion model based on Swin Transformer. Particularly, we design feature selection gates, which are integrated with the feature extraction layers to detect outliers and block them from HDR image synthesis. To reconstruct the missing local details by well-aligned and properly-exposed regions, we exploit the long distance contextual dependency in the exposure-space pyramid by the self-attention mechanism. Extensive numerical and visual evaluation has been conducted on a variety of benchmark datasets. The experiments show that our model achieves the accuracy on par with current top performing multi-exposure HDR imaging models, while gaining higher efficiency.
  
\end{abstract}

\section{Introduction}
The dynamic range of current consumer-level light sensors is still quite limited. As a result, a single shot of the natural scene always fails to capture image details in the strongly or weakly lit scene areas. A promising approach is to synthesize a high dynamic range image by a collection of images captured at multiple exposures, with the over-exposed or under-exposed of one image being well captured in other images \cite{SIGGRAPH97,ward2003fast}. This approach is challenged, if the camera and scene objects move independently. In such situations, to infer the missing contents of one exposure from other exposures, perfect motion estimation and sequence registration are required. Unfortunately, image motion estimation is an ongoing problem far from solved, being especially sensitive to brightness variation and lack of details. The erroneous estimation of motion would further cause the notorious ``ghosting'' or ``motion blur'' artifacts. Therefore, the central problem of multi-exposure HDR fusion is to fuse images containing large motion.

Many previous works detect misalignment by a post-processing step, then remove the detected areas from HDR image computation. The detectors are manually designed according to domain knowledge based on local  low-level features, such as appearance \cite{grosch2006fast}, gradient \cite{zhang2011gradient}, matrix rank \cite{lee2014ghost,oh2014robust}, low-order statistics \cite{pece2010bitmap}. The representation power of such features is limited, especially in under-exposed or over-exposed regions. Consequently, they may fail to locate well-exposed correspondences in side images to ill-posed regions in the reference image.    

The application of Convolutional Neural Networks (CNN) to HDR imaging enables extracting high-level features to guide multi-exposure fusion \cite{Kalantari2017DeepHD}. Joint alignment and HDR fusion models have been shown effective \cite{wu2018end,niu2021hdrgan}. In general, CNN-based multi-exposure HDR models focus on handling two issues: to establish long-range correlation and to suppress outlier features. To allow long-range image content retrieval, a large receptive field is necessary, hence 
dilated convolution \cite{yan2019attention}, deformable convolution \cite{liu2021adnet}, non-local module \cite{yan2020nonlocal} are explored. To distinguish inlier and outlier features, feature selection by spatial attention \cite{yan2019attention} and mask \cite{xiong2021hierarchical} are devised. Although the theoretical receptive field of a convolutional network may be expanded to the full image scale, the effective receptive field may still be local \cite{luo2016understanding}. 

Compared to CNN models, Transformer is more capable at global matching \cite{vaswani2017attention}. However, by the time of writing, its application to multi-exposure HDR fusion has not been discussed, to the best of our knowledge. The difficulty of applying Transformer is its high computational load, which causes imaging latency. A newly proposed technique Swin Transformer, whose shift-window self-attention progressively expands local attention to global attention, while gaining computational efficiency \cite{liu2021swin}. Nonetheless, directly stacking Swin Transformers for multi-exposure fusion consumes huge graphics memory and obstructs the network from extracting deep features. 

In this work, we investigate applying Swin Transformer to deep fusion of multi-exposure images. Inspired by Restormer \cite{zamir2021restormer}, which modifies Transformer to gated transformer and embeds it to a U-net for single image restoration, we alter the Multi-Layer Perceptron (MLP) of Swin Transformer to a gating network. In the framework of hierarchical encoder-decoder, Gated Swin Transformer can be stacked to form deep network, which progressively establishes global correlation and dynamically suppresses outliers.   
To summarize, the presented work
\begin{itemize}
\item constructs a new model based on Swin Transformer for multi-exposure HDR imaging. 
\item proposes integrating Gated Swin Transformer with  pyramidal encoder-decoder residual network. It takes advantage of the computational efficiency of Swin Transformer, while alleviating the heavy graphics memory consumption of Swin Transformer by hierarchical computation. 
\item achieves accuracy on par with state-of-the-art methods, while gaining higher efficiency. 
\end{itemize}
\section{Related Work}
The key to multi-exposure HDR imaging is to exploit cross-exposure correlation. Early works assume that the scene is static, and synthesize the HDR image by inferring the light energy map and camera response function from exposure bracketing \cite{SIGGRAPH97,mitsunaga1999radiometric,pal2004probability} or by a weighted fusion of different exposures \cite{ward2003fast,gelfand2010multi}. The fusion approach is now more widely investigated. The advent of CNN recasts the fusion in image domain to fusion in feature space. Recently, research effort on multi-exposure fusion has been mainly devoted to addressing cross-exposure registration and misalignment, to overcome the difficulty caused by scene motion \cite{ma2017robust}. 

\textbf{CNN-Based Multi-Exposure Fusion for HDR.} Applying deep convolution to multi-exposure fusion allows establishing cross-exposure correlation by extracting high-level contextual features, yielding robust matching. The first CNN-based model estimates image motion by optical flow to align the image sequence before convolutional feature extraction \cite{Kalantari2017DeepHD}. This external registration step prevents the model from being end-to-end trainable. Moreover, accurate optical flow algorithms are generally too slow for real-time HDR imaging. \cite{wu2018end} propose an end-to-end image translation network to formulate multi-image HDR fusion. Generative Adversarial Network is investigated, implicitly aligns the images in the generator \cite{niu2021hdrgan}. Several works perform soft feature selection. For example, \cite{yan2019attention} designs a spatial attention module by filtering the concatenation of low-level features extracted from the reference and side images, whereas \cite{xiong2021hierarchical} trains a masking tensor to blend the exposures. These models show that feature selection is crucial for HDR imaging. Recently, \cite{liu2021adnet} and \cite{xiong2021hierarchical} demonstrate the benefits of pyramidal alignment and fusion.

Different from the CNN-based models, our network explores the potential of Swin Transformer for HDR imaging. The self-attention mechanism in Swin Transformer implicitly searches for global cross-exposure alignment. We modify Swin Transformer with the gating mechanism similar to \cite{yu2019free} to mask out outliers.

\textbf{Transformer Network for Single Image Restoration}
A related line of work is single image restoration, e.g., denoising, deblurring, inpainting, super-resolution, based on modified Transformer (e.g., \cite{zamir2021restormer}) or Swin Transformer (e.g., \cite{liang2021swinir}).  

Our work is closely related to Restormer. Briefly, Restormer attaches a gating unit to the feed-forward network in Transformer. To lower the computational cost inherent to Transformer, Restormer replaces the pairwise affinity computation in Transformer by cross-channel covariance. That is, it uses channel attention instead of spatial attention, therefore makes it unsuitable for inferring cross-exposure spatial correlation. Instead, we employ Swin Transformer, which replaces global self-attention by shifted window self-attention. Swin Transformer trades memory storage for computational efficiency, maintaining the capability of connecting spatially distant yet contextually close features. Inspired by the hierarchical architectures in \cite{zamir2021restormer,xiong2021hierarchical,liu2021adnet}, our Gated Swin Transformer units serve in a U-shape pyramidal encoder-decoder, which significantly saves the memory storage. 
\section{Approach}
\subsection{Overall Workflow}
\begin{figure*}[!tbh]
\centering
\includegraphics[width=0.95\textwidth]{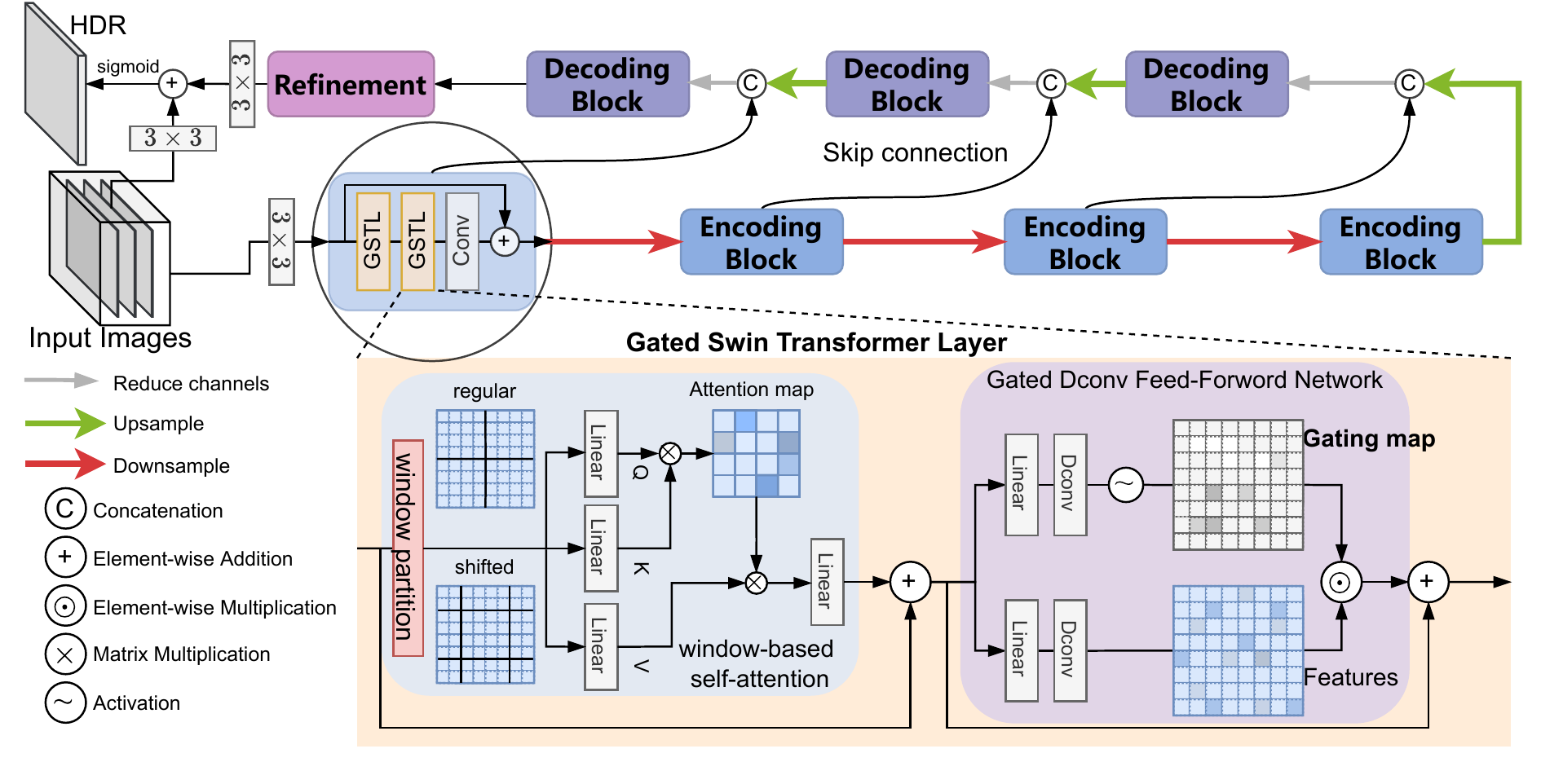}
\caption{Overview of our multi-frame HDR model. It is mainly composed of our proposed gated swin transformer layer. The enlarged part shows the internal structure of our proposed GSTL.}
\label{arch}
\end{figure*}
The multi-exposure fusion for HDR imaging can be described as, given an odd number of $K$ Limited Dynamic Range (LDR) images $\{\mathbf{I}_{k}\}_{k\in[1,K]}$, which are shot at different shutter length $t_{k\in[1,K]}$, where $t_1 < t_2 < \cdots < t_K$, transforming the middle exposure $\mathbf{I}_{\frac{k+1}{2}}$ to a HDR image $\hat{\mathbf{I}}$.  

Following the literature convention, this work assumes $K=3$ and normalizes $\mathbf{I}_{k}$ by $t_{k}$ according to
\begin{equation}
\mathbf{H}_{k}=\frac{\mathbf{I}_{k}^{\gamma}}{t_{k}},
\end{equation}
where by default $\gamma$ is 2.2. We denote the concatenation of $\mathbf{H}_{k}$ and $\mathbf{I}_{k}$ as $\mathbf{X}_{k}$, the concatenation of all $\mathbf{X}_{1\leq k \leq 3}$ as the input tensor $\mathbf{X}$, which is the input our neural network HDR model $f(\mathbf{X};\mathbf{\Theta})$, where $\mathbf{\Theta}$ stands for the set of trainable parameters. 

The architecture of $f(\mathbf{X};\mathbf{\Theta})$ is designed as a residual network. That is, a long shortcut connects the bottom level features with the top level features, to stabilize training. Mathematically, $f(\mathbf{X};\mathbf{\Theta})$ is decomposed to 
\begin{align}
    f(\mathbf{X};\mathbf{\Theta}) = 
    p\left(
    g\left(\mathbf{X}_{2};\mathbf{\Theta}_{g}\right)+
    r\left(\mathbf{X};\mathbf{\Theta}_{r}\right);
    \mathbf{\Theta}_{p}
    \right),
\end{align}
where $g\left(\mathbf{X}_{2};\mathbf{\Theta}_{g}\right)$ is a convolution layer that operates on $\mathbf{X}_{2}$; $p\left(~;\mathbf{\Theta}_{p}\right)$ consists of the final prediction convolution layer and a Sigmoid activation layer, to output $\hat{\mathbf{I}}$; $r\left(\mathbf{X};\mathbf{\Theta}_{r}\right)$ is the main body of the neural model that maps $\mathbf{X}$ to the residual. 

The residual function $r$ is designed as a U-shape Encoder-Decoder network. The encoder contains 4 encoding blocks $e_{1\leq i\leq 4}$, with a down-scaling layer between each pair of adjacent blocks, so as to extract hierarchical features. Different from Restormer \cite{zamir2021restormer}, each of these blocks is a small ResNet. Formally, the sequential encoding process can be descried as,
\begin{align}
    \mathbf{E}_{1} & =  e_{1}\left(\mathbf{X};\mathbf{\Theta}_{e_1}\right)
    +\mathbf{X} \\\nonumber
    \mathbf{E}_{i} & =  e_{i}\left(s\left(\mathbf{E}_{i-1}\right);
    \mathbf{\Theta}_{e_i}\right)+\mathbf{E}_{i-1}, \quad i>1,
    \label{eq:encode}
\end{align}
where $\mathbf{E}_{i}$ denote feature tensors; Function $s$ represents the down-scaling layer, implemented by Pixel-Unshuffle \cite{shi2016real} in our work. 

The decoder reverses the encoding hierarchy by 3 sequential decoding ResNet blocks $d_{i=3,2,1}$, with up-scaling layers inserted in between. The feature tensors are computed by
\begin{align}
    \mathbf{D}_{3} & =  d_{3}\left([u(\mathbf{E}_{3}),\mathbf{E}_{2}];\mathbf{\Theta}_{d_3}\right)+\mathbf{E}_{3}
    \\\nonumber
    \mathbf{D}_{i} & =  d_{i}\left([u(\mathbf{D}_{i+1}),\mathbf{E}_{i}];\mathbf{\Theta}_{d_i}\right)+\mathbf{D}_{i+1}, \quad i<3,
\end{align}
where function $u$ performs up-sampling by Pixel-Shuffle, Notation $[\cdot,\cdot]$ means channel-wise concatenation. 

The decoded feature tensor $\mathbf{D}_{1}$ is further processed by a refinement block $t$ (see Sec. \ref{sec:block}). The overall architecture of the proposed model is depicted in Fig. \ref{arch}.

\subsection{Internal Architecture of Building Blocks}
\label{sec:block}
The encoding blocks $e_{1\leq i\leq 4}$, decoding blocks $d_{1\leq i\leq 3}$, and refinement block $t$ have similar internal architecture. For instance, encoder $e_{i}$ processes the input tensor $\mathbf{E}_{i-1}$ (Eq. \ref{eq:encode}) by $n_{i}$ sequential Residual Gated Swin Transformers (RGST, detailed in Sec. \ref{sec:Attention}), followed by a convolution layer. Then a skip connection sums up the convolution output and $\mathbf{E}_{i-1}$. The number of RGST units used in each block varies with block feature spatial resolution. Here we set the numbers for the encoding and decoding blocks to $[2,3,3,4]$, from the fine to coarse scale in the feature pyramids. The number of the RGST units in the refinement block is set to $2$. 

\subsection{Residual Gated Swin Transformer Unit}
\label{sec:Attention}
Our basic building unit RGST alters Swin Transformer to address the multi-exposure HDR fusion problem. We take advantage of the multi-head self-attention modules in Swin Transformer, to aggregate features according to pair-wise feature affinity, which implicitly performs cross-exposure alignment and feature fusion. Furthermore, we modify the Multi-Perception Layer (MPL) in Swin Transformer to a gated feed-forward network, where the gates are defined according to \cite{zamir2021restormer}.

\subsubsection{Self-Attention Computation}
We take the $i$-th encoding block $e_i$ as an example to analyze the process of feature aggregation weighted by windowed affinity in the RGST unit. The input feature tensor $\mathbf{E}_{i}$ is layer-wisely normalized to $\mathbf{E}_{i}$, and then partitioned to windows of size $M\times M$. In each window, $\mathbf{E}_{i}$ is linearly projected to generate the Query, Key and Value tensors by trainable matrices $\mathbf{W}_q^\mathrm{j}$, $\mathbf{W}_k^\mathrm{j}$, $\mathbf{W}_v^\mathrm{j}$, where $j$ indexes the sets of projections used in the attention module, each set counted as one ``head'' in the literature. Local pair-wise affinity is computed within each window, by multiplying the Query and Key matrices. Subsequently, each feature in the Vector matrix is replaced by an aggregation of local features weighted by the computed affinity. The output of each head are concatenated and the linearly projected to a tensor ${\mathbf{E}^{\prime}_{i}}$. Unlike global self-attention, this windowed self-attention compares each feature against locally neighbouring features. Hence it is prone to misalignment caused by large motion, which is the so-called ``aperture problem'' in the field of optical flow computation. As a result, the aggregation may be impacted heavily by misaligned  features (i.e., outliers). We refer readers to \cite{liu2021swin} for details of Swin Transformer.

\subsubsection{Gated Feed-Forward Network}
To restrain the contaminated features from further computation, we employ the gating mechanism proposed by \cite{zamir2021restormer}. In particular, ${\mathbf{E}^{\prime}_{i}}$ is first layer-wisely normalized to ${\mathbf{E}^{\prime}}_{i}$. It then enters two parallel convolution branches, each of which is a pair of pixel-wise and channel-wise convolutions, yielding tensors $\dot{\mathbf{E}}_{i}$ and $\ddot{\mathbf{E}}_{i}$. $\dot{\mathbf{E}}_{i}$ is further activated by the Gaussian Error Linearity Unit (GELU) function \cite{hendrycks2016gaussian}, obtaining the gate tensor $\mathbf{G}_{i}$, which is point-wisely multiplied to $\ddot{\mathbf{E}}_{i}$, yielding $\bar{\mathbf{E}}_{i}$. 

The rest procedure inside the encoding block $e_{i}$ involves the self-attention in shifted windows and another gated feed-forward network. Their computation is as same as the description above. Although the window shifting strategy enlarges the receptive field, it may not be fully global at the finest scale of the U-net hierarchy. As the computation proceeds to coarser scales, the receptive field is significantly enlarged to the full image. Hence using the Gated Swin Transformer as the building unit in the U-shape encoder-decoder network achieves both global receptive field and low memory consumption. 

In our work, the window width $M$ is set to 8. The number of heads for self-attention varies with the feature spatial resolution, taking the value of \([1,2,4,4]\) from fine to coarse.

\subsection{Loss Function}
The loss function to train our model consists of two loss terms. Following the literature, the first loss term $\mathcal{L}_1$ measures the closeness of the output $\hat{\mathbf{I}}$ and the ground truth $\mathbf{I}$ in the tone-mapped domain by $l_{1}$ norm distance
\begin{equation}
\mathcal{L}_1=\|\mathcal{T}(\mathbf{I})-\mathcal{T}(\hat{\mathbf{I}})\|_{1} 
\end{equation}
where the tone mapping function $\mathcal{T}$ is the differentiable $\mu$-law range compressor, commonly used in previous works \cite{Kalantari2017DeepHD}.

Additionally, to train the model to faithfully reconstruct the missing details in over-exposed and under-exposed regions, the second loss term $\mathcal{L}_2$ measures the Structural Similarity (SSIM) \cite{wang2004image}, between the tone-mapped output and ground truth,
\begin{equation}
\mathcal{L}_{2}=1-\text{SSIM}(\mathcal{T}(\mathbf{I}), \mathcal{T}(\hat{\mathbf{I}})) 
\end{equation}

The total loss function $\mathcal{L}$ is then defined by 
\begin{equation}
\mathcal{L} = \mathcal{L}_{1}+\mathcal{L}_{2}.
\end{equation}

\section{Experimental Results}
\label{sec:experiments}
\subsection{Training Details}
\cite{Kalantari2017DeepHD} provides the first dataset specifically designed for multi-exposure HDR fusion under large motion. It consists of 74 training sets, which we use to supervise the training of our model. We crop the input images to patches of size \(256 \times 256\) at a step size of 64. This totally generates 20128 training samples. To augment training samples, we randomly rotate and flip the training images. The training adopts Adam optimizer. The learning rate is initialized to \(10^{-4}\) and is reduced to \(10^{-5}\) after 20 epochs. It is observed that 40 epochs are sufficient for the training to converge.    

\subsection{Numerical Evaluation}
We numerically measure the performance of our method on the 15 test sets of \cite{Kalantari2017DeepHD}, by Peak Signal-to-Noise Ratio (PSNR) and Structure Similarity, computed in both tonemapping domain (-\(\mu\)) and HDR linear domain (-L). Visual difference metric HDR-VDP-2 is also adopted, where the parameters are set as same as in previous works \cite{wu2018end} and \cite{niu2021hdrgan}. 

Table \ref{table_metrics} compares our model with state-of-the-art models. For \cite{yan2020nonlocal} and \cite{xiong2021hierarchical}, we use the results reported in the publications. Note that \cite{sen2012robust} and \cite{hu2013hdr} are not machine learning based methods. Moreover,  \cite{Kalantari2017DeepHD} and \cite{wu2018end} apply optical flow and homography transformation to preprocess the input images respectively, and hence entail extra computation. 

Table \ref{table_metrics} shows that our method outperforms competing method in terms of PSNR-L, SSIM-$\mu$, SSIM-L and HDR-VDP-2. It ranks the second best in PSNR-$\mu$, being slightly (0.1dB) inferior to \cite{xiong2021hierarchical}. Note that \cite{xiong2021hierarchical} utilizes a pretrained model to detect ghosting regions for training, whereas our method does not require any pretrained model. The high PSNR and SSIM scores varify that our model has strong HDR reconstruction ability and can accurately restore the radiance and structure of the scene in both tonemapping domain and HDR linear domain. Furthermore, its high performance in term of HDR-VDP-2\cite{mantiuk2011hdr} performance indicates that our method can generate HDR image visually close to the target image.

\begin{table*}[ht]
\centering
\begin{tabular}{l|c|c|c|c|c}
\hline
& PSNR-$\mu$ & PSNR-L & SSIM-$\mu$ & SSIM-L & HDR-VDP-2 \\
\hline
\bfseries Sen & 40.97 & 38.36 & 0.9830 & 0.9746 & 60.60\\
\hline
\bfseries Hu  & 35.65 & 30.80 & 0.9725 & 0.9491 & 58.34\\
\hline
\bfseries Kalantari & 42.69 & 41.22 & 0.9888 & 0.9845 & 65.05\\
\hline
\bfseries DeepHDR& 41.99 & 41.22 & 0.9878 & 0.9859 & \underline{65.91}\\
\hline
\bfseries AHDR & 43.62 & 41.03 & 0.9900  &\underline{0.9883} & 63.85 \\
\hline 
\bfseries NHDRRNet& 42.414 & - & 0.9887 & - & 61.21 \\
\hline 
\bfseries HDR-GAN &43.92 & \underline{41.57} &\underline{0.9905} &0.9865 & 65.45\\
\hline 
\bfseries HFNet & \textbf{44.28} & 41.47 & - & - & - \\
\hline 
\bfseries Ours & \underline{44.18} & \textbf{42.19}&\textbf{0.9912} & \textbf{0.9883}& \textbf{67.07} \\
\hline
\end{tabular}
\caption{Numerical performance of the proposed model, evaluated on the dataset by Kalantari-Ramamoorthi. The best and second best results for each metric are marked in \textbf{bold} and \underline{underlined}, respectively}
\label{table_metrics}
\end{table*}

\subsection{Visual Performance Evaluation}

\begin{figure*}[!htb]
\centering
\includegraphics[width=\textwidth]{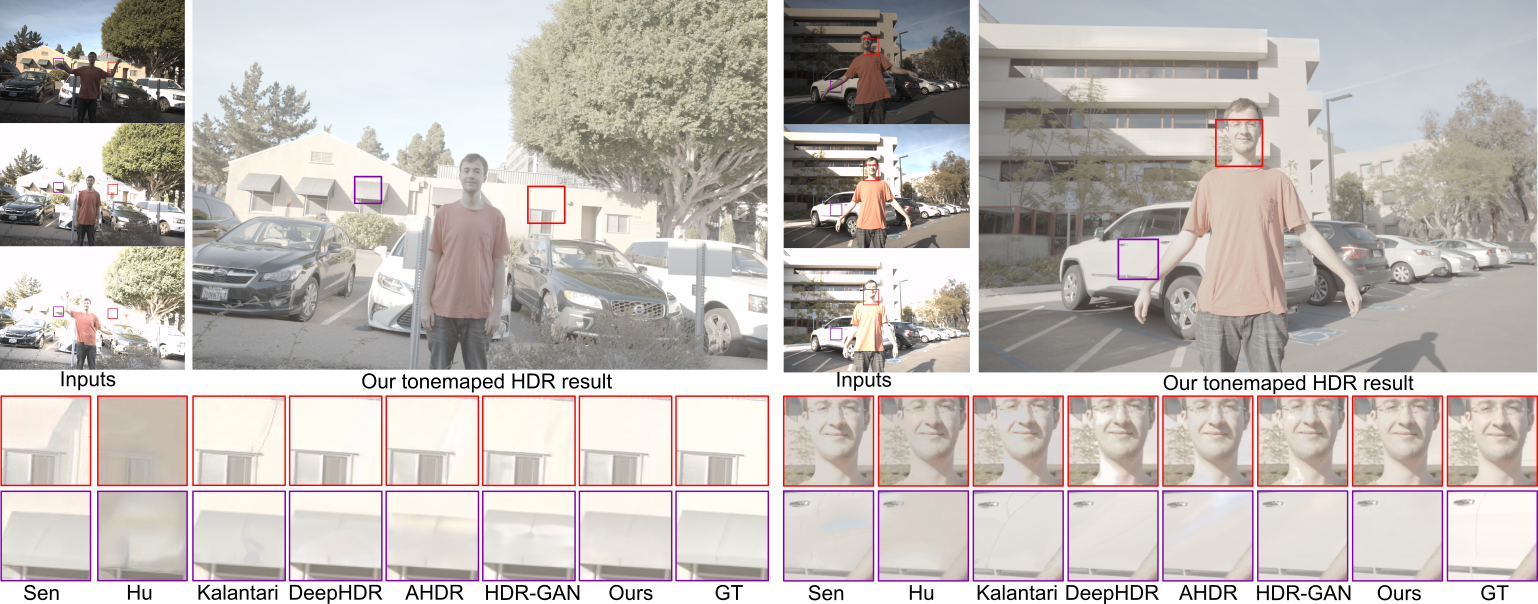}
\caption{Visual comparison on the test set of Kalantari-Ramamoorthi dataset. Zoom-in views of reconstruction by each method are presented on the saturated regions that contain moving objects. Our network built with gated Swin Transformer yields noticeably better visual results than other methods.}
\label{fig_kalantari_test}
\end{figure*}
Fig. \ref{fig_kalantari_test} present the visual performance of our method and comparable methods on two examples from \cite{Kalantari2017DeepHD}. We present the zoom-in views of two challenging cases, where large saturated regions contain substantial non-rigid motion in the reference image. The two patch-based methods do not reconstruct the missing details in the saturated regions, as they heavily rely on the details provided by the reference image for registration. Image reconstructed by the optical flow based method \cite{Kalantari2017DeepHD} suffers motion blur artifacts. This is because the convolutions of DeepHDR and HDR-GAN have limited receptive fields, and hence are hampered to repair missing content in misaligned regions by aligned regions. The gating mechanism of AHDR is only applied to low-level features, so the high-level outliers may deteriorate the HDR fusion. In contrast to comparable methods, our model remarkably overcomes the ghosting artifacts.

\begin{figure}[ht]
\centering
\includegraphics[width=\columnwidth]{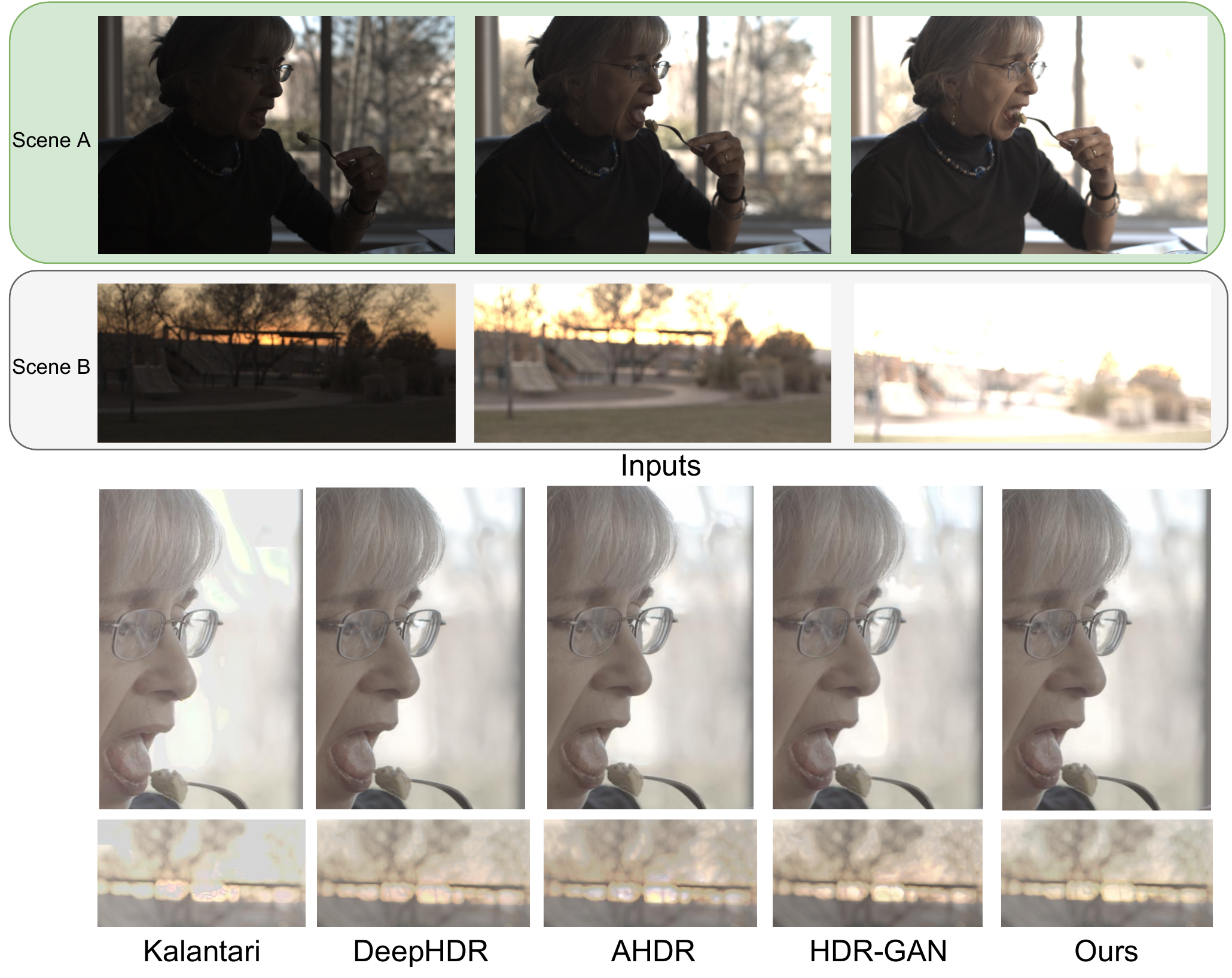}
\caption{Visual performance comparison on example images from the dataset by Sen et al. Zoom in views on challenging areas are presented. Although the ground truth is unavailable, it can be clearly observed that our method visually performs better than comparable methods.}
\label{sen_test}
\end{figure}

\begin{figure}[ht]
\centering
\includegraphics[width=\columnwidth]{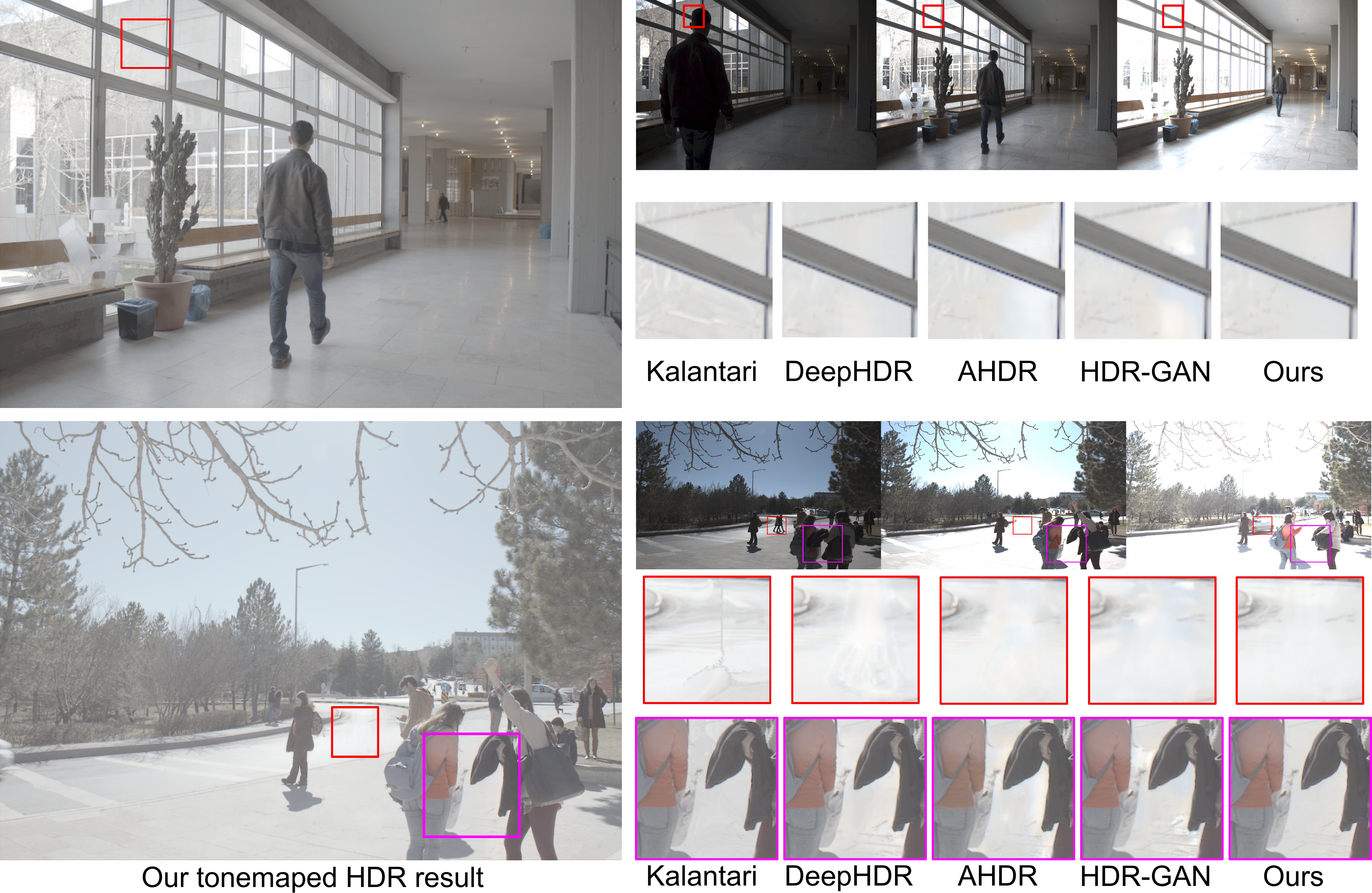}
\caption{Visual performance comparison on example images from the dataset by Tursun et al. Compared to state of the art methods, our method suffers less ghosting artifact.}
\label{tursun_test}
\end{figure}

Fig.\ref{sen_test} and Fig.\ref{tursun_test} present visual performance of our method on two examples from benchmark datasets \cite{sen2012robust} and \cite{tursun2016objective}. As these test datasets   do not provide ground truth image. we mark the visual difference on the results generated by different methods. It can be seen that our method suffers less artifacts than other methods in various scenes with various motion patterns, achieving better visual results. Our method creates high-quality HDR more robustly and generalizes well. 

\subsection{Ablation Study}

\begin{table}[h]
\centering
\resizebox{\columnwidth}{!}{
\begin{tabular}{l|c|c|c|c|c}
\hline
                         & PSNR-$\mu$ & PSNR-l & SSIM-$\mu$ & SSIM-l & HDR-VDP-2 \\ \hline
restormer(w/o ssim loss) & 44.00  & 41.5   & 0.9906 & 0.9873 & 64.72  \\ \hline
Ours(w/o ssim loss)      & 44.07  & 41.83  & 0.9909 & 0.9879 &  64.78  \\ \hline
Ours                     & 44.18  & 42.19  & 0.9912 & 0.9883 & 67.07      \\ \hline
\end{tabular}
}
\caption{Experimental results of ablation study. We compare using Gated Swin Transformer v.s. Gated Transformer, and the combined loss function v.s. the traditional $l_{1}$ norm loss function.}
\label{table_ablation_block_loss}
\end{table}

We verify various components of our method, including Swin Transformer, loss function, and gating mechanism by ablation study.

\subsubsection{Ablation Study on Block Design}
Our model has similar architecture to Restormer, which uses modified Transformer, whereas we use modified Swin Transformer as the building unit. For comparison, we replace the residual modules in each block in our model with multiple transformer layers as in Restormer, with same number of transformer layers. Table \ref{table_ablation_block_loss} presents the results, which show that using Swin Transformer achieves superior performance in all measures. The reason is that the attention module of Restormer is computed channel-wise, but forgoes the cross-exposure spatial dependency to repair the non-aligned area. 

\subsubsection{Ablation Study on Loss Function}
We trained our model under different loss function configurations, as shown in \ref{table_ablation_block_loss}. The results validate that the SSIM loss benefits detail reconstruction.

\subsubsection{Ablation Study on Gating Mechanism}
\begin{table}[h]
\resizebox{\columnwidth}{!}{
\begin{tabular}{l|c|c|c|c|c}
\hline
           & PSNR-$\mu$ & PSNR-l & SSIM-$\mu$ & SSIM-l & HDR-VDP-2 \\ \hline
w/o gating & 43.14  & 41.03  & 0.9904 & 0.9868 &     64.88      \\ \hline
one gating & 43.44  & 41.42  & 0.9909 & 0.9882 &    67.13   \\ \hline
Ours       & 43.61  & 41.74  & 0.9909 & 0.9881 & 66.96     \\ \hline
\end{tabular}
}
\caption{Ablation experimental results to verify the effectiveness of the gating mechanism}
\label{table_ablation_gating}
\end{table}

The gating mechanism is an important component in our model. Ablation study is conducted in the gating mechanism as follows.

\textbf{w/o gating}: The gating mechanism is not used in the feed forward network of all transformer layers in the model, that it, our GST unit degenerate to the vanilla Swin Transformer.

\textbf{one gating}: The gating mechanism is only used in the first Swin Transformer layers subsequent to the embedding layer, but not used for other layers. 

 Table \ref{table_ablation_gating} shows the results of the ablation experiments, where the model is trained for 20 epochs. By removing the gating mechanism, the network relies on self-attention for image alignment, resulting in the lowest performance. On top of it, adding gates to low level layers notably improves the HDR reconstruction. Furthermore, by integrating the gating mechanism with all Swin Transformer layers, the model effectively inpaints information in non-aligned regions and obtains the highest HDR reconstruction results, thus validates the effectiveness of the gating mechanism in our model.

\section{Conclusion}
High Dynamic Range (HDR) Imaging is a desirable function for modern digital cameras. To overcome the hardware limitation of current camera sensors in dynamic range, we have described a new neural model to blend low dynamic range images captured at tri-exposures into HDR images. We have shown that, by integrating the gating mechanism with the shifted-window self-attention mechanism, our model progressively aligns the cross-exposure images and screen outliers caused by large motion from HDR fusion. We have also demonstrated that, in the framework of multi-scale encoder-decoder, the Gated Swin Transformer achieves global reception, long-range matching and adaptive fusion. Extensive ablation study has been presented to validate the effectiveness of our model in addressing large motion in HDR imaging.

\bibliographystyle{named}
\bibliography{ijcai22}

\begin{thebibliography}{}

\bibitem[\protect\citeauthoryear{Debevec and Malik}{1997}]{SIGGRAPH97}
Paul~E. Debevec and Jitendra Malik.
\newblock Recovering high dynamic range radiance maps from photographs.
\newblock In {\em SIGGRAPH}, page 369–378, 1997.

\bibitem[\protect\citeauthoryear{Gelfand \bgroup \em et al.\egroup
  }{2010}]{gelfand2010multi}
Natasha Gelfand, Andrew Adams, Sung~Hee Park, and Kari Pulli.
\newblock Multi-exposure imaging on mobile devices.
\newblock In {\em ACM international conference on Multimedia}, pages 823--826,
  2010.

\bibitem[\protect\citeauthoryear{Grosch and others}{2006}]{grosch2006fast}
Thorsten Grosch et~al.
\newblock Fast and robust high dynamic range image generation with camera and
  object movement.
\newblock {\em Vision, Modeling and Visualization, RWTH Aachen}, 277284, 2006.

\bibitem[\protect\citeauthoryear{Hendrycks and
  Gimpel}{2016}]{hendrycks2016gaussian}
Dan Hendrycks and Kevin Gimpel.
\newblock Gaussian error linear units (gelus).
\newblock {\em arXiv:1606.08415}, 2016.

\bibitem[\protect\citeauthoryear{Hu \bgroup \em et al.\egroup
  }{2013}]{hu2013hdr}
Jun Hu, Orazio Gallo, Kari Pulli, and Xiaobai Sun.
\newblock Hdr deghosting: How to deal with saturation?
\newblock In {\em Computer Vision and Pattern Recognition}, pages 1163--1170,
  2013.

\bibitem[\protect\citeauthoryear{Kalantari and
  Ramamoorthi}{2017}]{Kalantari2017DeepHD}
Nima~Khademi Kalantari and Ravi Ramamoorthi.
\newblock Deep high dynamic range imaging of dynamic scenes.
\newblock {\em ACM Transactions on Graphics}, 36:144, 2017.

\bibitem[\protect\citeauthoryear{Lee \bgroup \em et al.\egroup
  }{2014}]{lee2014ghost}
Chul Lee, Yuelong Li, and Vishal Monga.
\newblock Ghost-free high dynamic range imaging via rank minimization.
\newblock {\em IEEE signal processing letters}, 21(9):1045--1049, 2014.

\bibitem[\protect\citeauthoryear{Liang \bgroup \em et al.\egroup
  }{2021}]{liang2021swinir}
Jingyun Liang, Jiezhang Cao, Guolei Sun, Kai Zhang, Luc Van~Gool, and Radu
  Timofte.
\newblock Swinir: Image restoration using swin transformer.
\newblock In {\em International Conference on Computer Vision}, pages
  1833--1844, 2021.

\bibitem[\protect\citeauthoryear{Liu \bgroup \em et al.\egroup
  }{2021a}]{liu2021swin}
Ze~Liu, Yutong Lin, Yue Cao, Han Hu, Yixuan Wei, Zheng Zhang, Stephen Lin, and
  Baining Guo.
\newblock Swin transformer: Hierarchical vision transformer using shifted
  windows.
\newblock {\em arXiv preprint arXiv:2103.14030}, 2021.

\bibitem[\protect\citeauthoryear{Liu \bgroup \em et al.\egroup
  }{2021b}]{liu2021adnet}
Zhen Liu, Wenjie Lin, Xinpeng Li, Qing Rao, Ting Jiang, Mingyan Han, Haoqiang
  Fan, Jian Sun, and Shuaicheng Liu.
\newblock Adnet: Attention-guided deformable convolutional network for high
  dynamic range imaging.
\newblock In {\em CVPR}, pages 463--470, 2021.

\bibitem[\protect\citeauthoryear{Luo \bgroup \em et al.\egroup
  }{2016}]{luo2016understanding}
Wenjie Luo, Yujia Li, Raquel Urtasun, and Richard Zemel.
\newblock Understanding the effective receptive field in deep convolutional
  neural networks.
\newblock In {\em NIPS}, pages 4905--4913, 2016.

\bibitem[\protect\citeauthoryear{Ma \bgroup \em et al.\egroup
  }{2017}]{ma2017robust}
Kede Ma, Hui Li, Hongwei Yong, Zhou Wang, Deyu Meng, and Lei Zhang.
\newblock Robust multi-exposure image fusion: a structural patch decomposition
  approach.
\newblock {\em IEEE Transactions on Image Processing}, 26(5):2519--2532, 2017.

\bibitem[\protect\citeauthoryear{Mantiuk \bgroup \em et al.\egroup
  }{2011}]{mantiuk2011hdr}
Rafa{\l} Mantiuk, Kil~Joong Kim, Allan~G Rempel, and Wolfgang Heidrich.
\newblock Hdr-vdp-2: A calibrated visual metric for visibility and quality
  predictions in all luminance conditions.
\newblock {\em ACM Transactions on graphics (TOG)}, 30(4):1--14, 2011.

\bibitem[\protect\citeauthoryear{Mitsunaga and
  Nayar}{1999}]{mitsunaga1999radiometric}
Tomoo Mitsunaga and Shree~K Nayar.
\newblock Radiometric self calibration.
\newblock In {\em conference on computer vision and pattern recognition}, pages
  374--380, 1999.

\bibitem[\protect\citeauthoryear{Niu \bgroup \em et al.\egroup
  }{2021}]{niu2021hdrgan}
Yuzhen Niu, Jianbin Wu, Wenxi Liu, Wenzhong Guo, and Rynson~WH Lau.
\newblock Hdr-gan: Hdr image reconstruction from multi-exposed ldr images with
  large motions.
\newblock {\em IEEE Transactions on Image Processing}, 30:3885--3896, 2021.

\bibitem[\protect\citeauthoryear{Oh \bgroup \em et al.\egroup
  }{2014}]{oh2014robust}
Tae-Hyun Oh, Joon-Young Lee, Yu-Wing Tai, and In~So Kweon.
\newblock Robust high dynamic range imaging by rank minimization.
\newblock {\em IEEE transactions on pattern analysis and machine intelligence},
  37(6):1219--1232, 2014.

\bibitem[\protect\citeauthoryear{Pal \bgroup \em et al.\egroup
  }{2004}]{pal2004probability}
Chris Pal, Richard Szeliski, Matthew Uyttendaele, and Nebojsa Jojic.
\newblock Probability models for high dynamic range imaging.
\newblock In {\em Conference on Computer Vision and Pattern Recognition}, pages
  173--180, 2004.

\bibitem[\protect\citeauthoryear{Pece and Kautz}{2010}]{pece2010bitmap}
Fabrizio Pece and Jan Kautz.
\newblock Bitmap movement detection: Hdr for dynamic scenes.
\newblock In {\em Conference on Visual Media Production}, pages 1--8. IEEE,
  2010.

\bibitem[\protect\citeauthoryear{Sen \bgroup \em et al.\egroup
  }{2012}]{sen2012robust}
Pradeep Sen, Nima~Khademi Kalantari, Maziar Yaesoubi, Soheil Darabi, Dan~B
  Goldman, and Eli Shechtman.
\newblock Robust patch-based hdr reconstruction of dynamic scenes.
\newblock {\em ACM Trans. Graph.}, 31(6):203--1, 2012.

\bibitem[\protect\citeauthoryear{Shi \bgroup \em et al.\egroup
  }{2016}]{shi2016real}
Wenzhe Shi, Jose Caballero, Ferenc Husz{\'a}r, Johannes Totz, Andrew~P Aitken,
  Rob Bishop, Daniel Rueckert, and Zehan Wang.
\newblock Real-time single image and video super-resolution using an efficient
  sub-pixel convolutional neural network.
\newblock In {\em Proceedings of the IEEE conference on computer vision and
  pattern recognition}, pages 1874--1883, 2016.

\bibitem[\protect\citeauthoryear{Tursun \bgroup \em et al.\egroup
  }{2016}]{tursun2016objective}
Okan~Tarhan Tursun, Ahmet~O{\u{g}}uz Aky{\"u}z, Aykut Erdem, and Erkut Erdem.
\newblock An objective deghosting quality metric for hdr images.
\newblock In {\em Computer Graphics Forum}, volume~35, pages 139--152. Wiley
  Online Library, 2016.

\bibitem[\protect\citeauthoryear{Vaswani \bgroup \em et al.\egroup
  }{2017}]{vaswani2017attention}
Ashish Vaswani, Noam Shazeer, Niki Parmar, Jakob Uszkoreit, Llion Jones,
  Aidan~N Gomez, {\L}ukasz Kaiser, and Illia Polosukhin.
\newblock Attention is all you need.
\newblock In {\em Advances in neural information processing systems}, pages
  5998--6008, 2017.

\bibitem[\protect\citeauthoryear{Wang \bgroup \em et al.\egroup
  }{2004}]{wang2004image}
Zhou Wang, Alan~C Bovik, Hamid~R Sheikh, and Eero~P Simoncelli.
\newblock Image quality assessment: from error visibility to structural
  similarity.
\newblock {\em IEEE transactions on image processing}, 13(4):600--612, 2004.

\bibitem[\protect\citeauthoryear{Ward}{2003}]{ward2003fast}
Greg Ward.
\newblock Fast, robust image registration for compositing high dynamic range
  photographs from hand-held exposures.
\newblock {\em Journal of graphics tools}, 8(2):17--30, 2003.

\bibitem[\protect\citeauthoryear{Wu \bgroup \em et al.\egroup
  }{2018}]{wu2018end}
Shangzhe Wu, Jiarui Xu, Yu-Wing Tai, and Chi-Keung Tang.
\newblock End-to-end deep hdr imaging with large foreground motions.
\newblock In {\em ECCV}, 2018.

\bibitem[\protect\citeauthoryear{Xiong and Chen}{2021}]{xiong2021hierarchical}
Pengfei Xiong and Yu~Chen.
\newblock Hierarchical fusion for practical ghost-free high dynamic range
  imaging.
\newblock In {\em ACM International Conference on Multimedia}, pages
  4025--4033, 2021.

\bibitem[\protect\citeauthoryear{Yan \bgroup \em et al.\egroup
  }{2019}]{yan2019attention}
Qingsen Yan, Dong Gong, Qinfeng Shi, Anton van~den Hengel, Chunhua Shen, Ian
  Reid, and Yanning Zhang.
\newblock Attention-guided network for ghost-free high dynamic range imaging.
\newblock In {\em Computer Vision and Pattern Recognition}, pages 1751--1760,
  2019.

\bibitem[\protect\citeauthoryear{Yan \bgroup \em et al.\egroup
  }{2020}]{yan2020nonlocal}
Qingsen Yan, Lei Zhang, Yu~Liu, Yu~Zhu, Jinqiu Sun, Qinfeng Shi, and Yanning
  Zhang.
\newblock Deep hdr imaging via a non-local network.
\newblock {\em IEEE Transactions on Image Processing}, 29:4308--4322, 2020.

\bibitem[\protect\citeauthoryear{Yu \bgroup \em et al.\egroup
  }{2019}]{yu2019free}
Jiahui Yu, Zhe Lin, Jimei Yang, Xiaohui Shen, Xin Lu, and Thomas~S Huang.
\newblock Free-form image inpainting with gated convolution.
\newblock In {\em Proceedings of the IEEE/CVF International Conference on
  Computer Vision}, pages 4471--4480, 2019.

\bibitem[\protect\citeauthoryear{Zamir \bgroup \em et al.\egroup
  }{2021}]{zamir2021restormer}
Syed~Waqas Zamir, Aditya Arora, Salman Khan, Munawar Hayat, Fahad~Shahbaz Khan,
  and Ming-Hsuan Yang.
\newblock Restormer: Efficient transformer for high-resolution image
  restoration.
\newblock {\em arXiv preprint arXiv:2111.09881}, 2021.

\bibitem[\protect\citeauthoryear{Zhang and Cham}{2011}]{zhang2011gradient}
Wei Zhang and Wai-Kuen Cham.
\newblock Gradient-directed multiexposure composition.
\newblock {\em IEEE Transactions on Image Processing}, 21(4):2318--2323, 2011.

\end{thebibliography}

\end{document}